\documentclass[sigconf]{acmart}
\usepackage[utf8]{inputenc}
\usepackage{natbib}
\usepackage{algorithm}
\usepackage{algpseudocode}
\usepackage{booktabs}
\usepackage{enumitem}
\usepackage{url}

\usepackage{pgfplotstable}
\usetikzlibrary{pgfplots.groupplots}

\pgfplotsset{compat=1.14}
\pgfplotsset{
    title style={yshift=-1.ex}, 
    tick label style = {font = \tiny},
    label style = {font = \small},
    legend style = {font = \footnotesize},
    every axis plot/.append style = {font = \footnotesize},
    every mark/.append style={solid},
}

\usepackage{amsmath,amsfonts,bm}

\def\eqref#1{equation~\ref{#1}}
\def\Eqref#1{Equation~\ref{#1}}

\def\1{\bm{1}}

\def\vh{{\bm{h}}}

\def\vv{{\bm{v}}}

\def\vz{{\bm{z}}}

\def\mV{{\bm{V}}}

\def\mX{{\bm{X}}}

\def\mZ{{\bm{Z}}}

\DeclareMathAlphabet{\mathsfit}{\encodingdefault}{\sfdefault}{m}{sl}
\SetMathAlphabet{\mathsfit}{bold}{\encodingdefault}{\sfdefault}{bx}{n}

\def\gA{{\mathcal{A}}}
\def\gB{{\mathcal{B}}}
\def\gC{{\mathcal{C}}}
\def\gD{{\mathcal{D}}}
\def\gE{{\mathcal{E}}}

\def\gS{{\mathcal{S}}}

\def\gV{{\mathcal{V}}}
\def\gW{{\mathcal{W}}}
\def\gX{{\mathcal{X}}}

\def\sR{{\mathbb{R}}}

\newcommand{\Ls}{\mathcal{L}}
\newcommand{\R}{\mathbb{R}}

\newcommand{\softmax}{\mathrm{softmax}}

\newcommand{\normlone}{L^1}

\DeclareMathOperator*{\argmax}{arg\,max}

\copyrightyear{2020}
\acmYear{2020}
\setcopyright{rightsretained}
\acmConference[CCS '20]{Proceedings of the 2020 ACM SIGSAC Conference on Computer and Communications Security}{November 9--13, 2020}{Virtual Event, USA}
\acmBooktitle{Proceedings of the 2020 ACM SIGSAC Conference on Computer and Communications Security (CCS '20), November 9--13, 2020, Virtual Event, USA}\acmDOI{10.1145/3372297.3417270}
\acmISBN{978-1-4503-7089-9/20/11}

\title{Information Leakage in Embedding Models}
\author{Congzheng Song}
\affiliation{Cornell University}
\email{cs2296@cornell.edu}
\authornote{Work done while at Google Brain.}
\author{Ananth Raghunathan}
\authornotemark[1]
\affiliation{Facebook}
\email{ananthr@cs.stanford.edu}

\newcommand{\enc}{\Phi}
\newcommand{\enci}{\Upsilon}
\newcommand{\emb}{\enc(x)}
\newcommand{\embt}{\enc(x^*)}

\newcommand{\vocab}{\gV}
\newcommand{\temp}{T}
\newcommand{\data}{\gD}
\newcommand{\datatarget}{\data_\textrm{target}}
\newcommand{\dataaux}{\data_{\textrm{aux}}}
\newcommand{\datatrain}{\data_{\textrm{train}}}
\newcommand{\etarget}{\gE_{\textrm{target}}}
\newcommand{\simscore}{\delta}
\newcommand{\threshsparse}{\tau_\textrm{sp}}
\newcommand{\threshmia}{\tau_m}

\newcommand{\paragraphb}[1]{\vspace{0.75ex}\noindent{\bf #1}\hspace*{.3em}}

\algtext*{EndWhile}
\algtext*{EndFor}
\algtext*{EndIf}
\algtext*{EndProcedure}
\begin{document}
\fancyhead{}

\begin{abstract}
Embeddings are functions that map raw input data to low-dimensional vector representations, while preserving important semantic information about the inputs. 
Pre-training embeddings on a large amount of unlabeled data and fine-tuning them for downstream tasks is now a de facto standard in achieving state of the art learning in many domains.

We demonstrate that embeddings, in addition  to encoding generic semantics, often also present a vector that leaks sensitive information about the input data.
We develop three classes of attacks to systematically study information that might be leaked by embeddings.
First, embedding vectors can be inverted to partially recover some of the input data. 
As an example, we show that our attacks on popular sentence embeddings recover between 50\%--70\% of the input words (F1 scores of 0.5--0.7). 
Second, embeddings may reveal sensitive attributes inherent in inputs and independent of the underlying semantic task at hand. 
Attributes such as authorship of text can be easily extracted by training an inference model on just a handful of labeled embedding vectors.
Third, embedding models leak moderate amount of membership information for infrequent training data  inputs.
We extensively evaluate our attacks on various state-of-the-art embedding models in the text domain. We also propose and evaluate defenses that can prevent the leakage to some extent at a minor cost in utility.

\end{abstract}


\maketitle

\section{Introduction}


Machine learning (ML) has seen an explosive growth over the past decade and is now widely used across industry from image analysis~\cite{krizhevsky2012imagenet,he2016deep}, speech recognition~\cite{hannun2014deep}, and even in applications in the medical sciences for diagnosis~\cite{choi2016doctor,rajkomar2018scalable}. These advances rely on not only improved training algorithms and architectures, but also access to high-quality and often sensitive data. The wide applications of machine learning and its reliance on quality training data necessitates a better understanding of how exactly ML models work and how they interact with their training data. Naturally, there is increasing literature focused on these aspects, both in the domains of interpretability and fairness of models, and their privacy. The latter is further of timely importance given recent regulations under the European Union's General Data Protection Regulation (GDPR) umbrella requiring users to have greater control of their data.

Applying a line of research investigating whether individual genomic records can be subsequently identified~\cite{homer2008,sankararaman2009genomic}, Shokri et al.~\cite{shokri2017membership} developed the first \emph{membership inference} tests investigating how ML models may leak some of their training data. 
This subsequently led to a large body of work exploring this space~\cite{nasr2018comprehensive,melis2019exploiting,yeom2018privacy,salem2018ml,sablayrolles2019white}. 
Another research direction investigating how models might memorize their training data has also shown promising results~\cite{carlini2019secret,song2019auditing}.
Apart from training data privacy, modern deep learning models can unintentionally leak information of the sensitive input from the model's representation at inference time~\cite{song2019overlearning,coavoux2018privacy,li2018towards,elazar2018adversarial,osia2018deep}.

This growing body of work analyzing models from the standpoint of privacy aims to answer natural questions: \emph{Besides the learning task at hand, what other information do models capture or expose about their training data?} And, \emph{what functionalities may be captured by ML models unintentionally?} We also note that beyond the scope of this paper are other rich sets of questions around adversarial behavior with ML models---the presence of adversarial examples, data poisoning attacks, and more.

\paragraphb{Embeddings.} In most of the research investigating the privacy of ML models, one class of models is largely conspicuously absent.
Embeddings are mathematical functions that map raw objects (such as words, sentences, images, user activities) to real valued vectors with the goal of capturing and preserving some important semantic meaning about the underlying objects. 
The most common application of embeddings is \emph{transfer learning} where the embeddings are pre-trained on a large amount of unlabeled raw data and later fine-tuned on downstream tasks with limited labeled data.
Transfer learning from embeddings have been shown to be tremendously useful for many natural language processing (NLP) tasks such as paraphrasing~\cite{dolan2005automatically}, response suggestion~\cite{kannan2016smart}, information retrieval, text classification, and question answering~\cite{rajpurkar2016squad}, where labeled data is typically expensive to collect.
Embeddings have also been successfully applied to other data domains including social networks~\cite{grover2016node2vec}, source code~\cite{alon2019code2vec}, YouTube watches~\cite{covington2016deep}, movie feedback~\cite{he2017neural}, locations~\cite{dadoun2019location}, etc. 
In some sense, their widespread use should not be surprising---embedding models and their success illustrate why deep learning has been successful at capturing interesting semantics from large quantities of raw data.

Embedding models are often pre-trained on raw and unlabeled data at hand, then used with labeled data to transfer learning to various downstream tasks.
Our study of embeddings is motivated by their widespread application and trying to better understand how these two stages of training may capture and subsequently leak information about sensitive data.
While it is attractive that these models inherently capture relations and similarities and other semantic relationships between raw objects like words or sentences, it also behooves one to consider if this is \emph{all} the information being captured by the embeddings. 
Do they, perhaps, in addition to meaning of words inadvertently capture information about the authors perhaps? Would that have consequences if these embeddings are used in adversarial ways? What aspects of the seminal research into privacy of ML models through the lens of membership inference attacks and memorization propensity can we apply to better understand the privacy risks of embeddings?

These questions and more lead us to initiate a systematic study of the privacy of embedding models by considering three broad classes of attacks. We first consider \emph{embedding inversion} attacks whose goal is to invert a given embedding back to the sensitive raw text inputs (such as private user messages). For embeddings susceptible to these attacks, it is imperative that we consider the embedding outputs containing inherently as much information with respect to risks of leakage as the underlying sensitive data itself. The fact that embeddings appear to be abstract real-numbered vectors should not be misconstrued as being safe.

Along the lines of Song and Shmatikov~\cite{song2019overlearning}, we consider \emph{attribute inference} attacks to test if embeddings might unintentionally reveal attributes of their input data that are sensitive; aspects that are not useful for their downstream applications. 
This threat model is particularly important when sensitive attributes orthogonal to downstream tasks are quickly revealed given very little \emph{auxiliary data}, something that the adversary might be able to capture through external means. As an example, given (sensitive) demographic information (such as gender) that an adversary might retrieve for a limited set of data points, the embeddings should ideally not enable even somewhat accurate estimates of these sensitive attributes across a larger population on which they might be used for completely orthogonal downstream tasks (e.g. sentiment analysis).

Finally, we also consider classic \emph{membership inference} attacks demonstrating the need to worry about leakage of training data membership when given access to embedding models and their outputs, as is common with many other ML models.

\paragraphb{Our Contributions.} As discussed above, we initiate a systematic and broad study of three classes of potential information leakage in embedding models. We consider word and sentence embedding models to show the viability of these attacks and demonstrate their usefulness in improving our understanding of privacy risks of embeddings. Additionally, we introduce new techniques to attacks models, in addition to drawing inspiration from existing attacks to apply them to embeddings to various degrees. Our results are demonstrated on widely-used word embeddings such as Word2Vec ~\cite{mikolov2013distributed}, FastText~\cite{bojanowski2017enriching}, GloVe~\cite{pennington2014glove} and different sentence embedding models including dual-encoders~\cite{logeswaran2018an} and BERT~\cite{devlin2019bert,lan2019albert}.

    \textbf{(1)} We demonstrate that sentence embedding vectors encode information about the exact words in input texts rather than merely abstract semantics.
    Under scenarios involving black-box and white-box access to embedding models, we develop several inversion techniques that can invert the embeddings back to the words with high precision and recall values (exceeding 60\% each) demonstrating that a significant fraction of the inputs may be easily recovered. We note that these techniques might be of independent interest.

    \textbf{(2)} We discover certain embeddings training frameworks in particular favor \emph{sensitive attributes} leakage by showing that such embedding models improve upon state-of-the-art stylometry techniques~\cite{ruder2016character,shetty2018a4nt} to infer text authorship with very little training data.
    Using dual encoder embeddings~\cite{logeswaran2018an} trained with contrastive learning framework~\cite{saunshi2019theoretical} and 10 to 50 labeled sentences per author, we show 1.5--3$\times$ fold improvement in classifying hundreds of authors over prior stylometry methods and embeddings trained with different learning paradigms.
    \smallskip

    \textbf{(3)} We show that membership inference attacks are still a viable threat for embedding models, albeit to a lesser extent. Given our results that demonstrate adversary can achieve a 30\% improvement on membership information over random guessing on both word and sentence embeddings, we show that is prudent to consider these potential avenues of leakage when dealing with embeddings.
    \smallskip

    \textbf{(4)} Finally, we propose and evaluate adversarial training techniques to minimize the information leakage via inversion and sensitive attribute inference. We demonstrate through experiments their applicability to mitigating these attacks. We show that embedding inversion attacks and attribute inference attacks against our adversarially trained model go down by 30\% and 80\% respectively. These come at a minor cost to utility indicating a mitigation that cannot simply be attributed to training poorer embeddings.
    \smallskip
    
\paragraphb{Paper Organization.} The rest of the paper is organized as follows. Section~\ref{sec:prelim} introduces preliminaries needed for the rest of the paper. Sections~\ref{sec:inversion}, ~\ref{sec:attribute}, and ~\ref{sec:membership} cover attacks against embedding models---inversion attacks, sensitive attributes inference attacks, and membership inference attacks respectively. Our experimental results and proposed defenses are covered in Sections~\ref{sec:expt} and~\ref{sec:defence} respectively, followed by related work and conclusions.


\section{Preliminaries}
\label{sec:prelim}
\subsection{Embedding Models}
\label{sec:prelim_emb}
Embedding models are widely used machine learning models that map a raw input (usually discrete) to a low-dimensional vector. 
The embedding vector captures the semantic meaning of the raw input data and can be used for various downstream tasks including nearest neighbor search, retrieval, and classification. 
In this work, we focus on embeddings of text input data as they are widely used in many applications and have been studied extensively in research community~\cite{kiros2015skip,pennington2014glove,mikolov2013distributed,bojanowski2017enriching,logeswaran2018an,cer2018universal,devlin2019bert,lan2019albert,radford2019language,reimers2019sentence}. 

\paragraphb{Word embeddings.} Word embeddings are look-up tables that map each word $w$ from a vocabulary $\vocab$ to a vector $\vv_w\in\sR^d$. 
Word embedding vectors capture the semantic meaning of words in the following manner: words with similar meanings will have small cosine distance in the embedding space, the cosine distance of $\vv_1$ and $\vv_2$ defined as $1-(\vv_1^\top\cdot\vv_2/\|\vv_1\|\|\vv_2\|)$.

Popular word embedding models including Word2Vec~\cite{mikolov2013distributed}, FastText~\cite{bojanowski2017enriching} and GloVe~\cite{pennington2014glove} are learned in an unsupervised fashion on a large unlabeled corpus.   
In detail, given a sliding window of words $\gC = [w_b, \dots, w_0, \dots, w_e]$ from the training corpus, Word2Vec and FastText train to predict the context word $w_{i}$ given the center word $w_0$ by maximizing the log-likelihood $\log P_\mV(w_i | w_0)$ where
\begin{align}
\label{eq:word}
    P_\mV(w_i | w_0) = \frac{\exp (\vv_{w_i}^\top\cdot\vv_{w_0})}
    {\sum_{w\in\{w_i\}\cup\gV_\textrm{neg}}\exp(\vv_{w}^\top\cdot\vv_{w_0})}
\end{align}
for each $w_i \in \gC / \{w_0\}$. To accelerate training, the above probability is calculated against $\gV_\textrm{neg}\subset\gV$, a sampled subset of words not in $\gC$ instead of the entire vocabulary. 
GloVe is trained to estimate the co-occurrence count of $w_i$ and $w_j$ for all pairs of $w_i, w_j\in \gC$.

A common practice in modern deep NLP models is to use word embeddings as the first layer so that discrete inputs are mapped to a continuous space and can be used for later computation.
Pre-trained word embeddings is often used to initialize the weights of the embedding layer. 
This is especially helpful when the downstream NLP tasks have a limited amount of labeled data as the knowledge learned by these pre-trained word embeddings from a large unlabeled corpus can be transferred to the downstream tasks.

\paragraphb{Sentence embeddings.} Sentence embeddings are functions that map a variable-length sequence of words $x$ to a fix-sized embedding vector $\emb\in\sR^d$ through a neural network model $\enc$.
For a input sequence of $\ell$ words $x = [w_1, w_2, \dots, w_\ell]$, $\enc$ first maps $x$ into a sequence of word vectors $\mX=[\vv_1, \dots, \vv_\ell]$ 
with a word embedding matrix $\mV$. 
Then $\enc$ feeds $\mX$ to a recurrent neural networks or Transformer~\cite{vaswani2017attention} and obtain a sequential hidden representation $[\vh_1, \vh_2, \dots, \vh_\ell]$ for each word in $x$.
Finally $\enc$ outputs the sentence embedding by reducing the sequential hidden representation to a single vector representation. Common reducing methods include taking the last representation where $\emb=\vh_\ell$ and mean-pooling where $\emb=(1/\ell) \cdot \sum_{i=1}^\ell\vh_i$.

Sentence embedding models are usually trained with unsupervised learning methods on a large unlabeled corpus. 
A popular architecture for unsupervised sentence embedding is the dual-encoder model proposed in many prior works~\cite{logeswaran2018an,cer2018universal,chidambaram2018learning,lowe2015ubuntu,henderson2017efficient,yang2018learning,reimers2019sentence}. 
The dual-encoder model trains on a pair of context sentences ($x_a, x_b$) where the pair could be a sentence and its next sentence in the same text or a dialog input and its response, etc. 
Given a randomly sampled set of negative sentences $\gX_\textrm{neg}$ that are not in the context of $x_a, x_b$, the objective of training is to maximizes the log-likelihood $\log P_\enc(x_b | x_a, \gX_\textrm{neg})$ where
\begin{align}
\label{eq:sent}
    P_\enc(x_b | x_a, \gX_\textrm{neg}) = \frac{\exp (\enc(x_b)^\top\cdot\enc(x_a))}
    {\sum_{x\in\{x_b\}\cup\gX_\textrm{neg}}\exp (\emb^\top\cdot\enc(x_a))}  .
\end{align}
Intuitively, the model is trained to predict the correct context $x_b$ from the set $\{x_b\}\cup\gX_\textrm{neg}$ when conditioned on $x_a$. 
In other words, the similarity between embeddings of context data is maximized while that between negative samples is minimized. 

Sentence embeddings usually outperform word embeddings on transfer learning. 
For downstream tasks such as image-sentence retrieval, classification, and paraphrase detection, sentence embeddings are much more efficient for the reason that only a linear model needs to be trained using embeddings as the feature vectors. 

\paragraphb{Pre-trained language models.} Language models are trained to learn contextual information by predicting next words in input text, and pre-trained language models can easily adopt to other NLP tasks by fine-tuning.
The recently proposed Transformer architecture~\cite{vaswani2017attention} enables language models to have dozen of layers with huge capacity.
These large language models, including BERT~\cite{devlin2019bert}, GPT-2~\cite{radford2019language}, and XLNet~\cite{yang2019xlnet}, are trained on enormous large corpus and have shown impressive performance gains when transferred to other downstream NLP tasks in comparison to previous state of the art methods.

These pre-trained language models can also be used to extract sentence embeddings.
There are many different ways to extract sentence-level features with pre-trained language models. 
In this work, we follow Sentence-BERT~\cite{reimers2019sentence} which suggests that mean-pooling on the hidden representations yields best empirical performance.

\begin{figure*}
    \centering
    \includegraphics[width=0.75\textwidth]{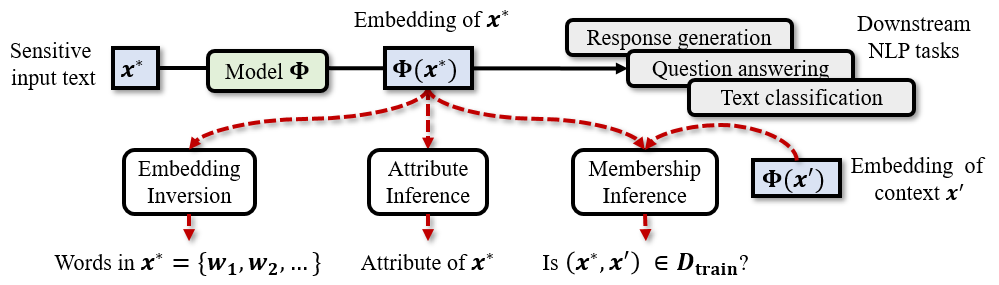}
    \caption{Taxonomy of attacks against embedding models. We assume adversary has access to the embedding $\enc(x^*)$ of a sensitive input text $x^*$ that will be used for downstream NLP tasks, and perform three information leakage attacks on $\enc(x^*)$: (1) inverting the embedding back to the exact words in $x^*$, (2) inferring sensitive attribute of $x^*$, and (3) inferring the membership, i.e. whether $x^*$ and its context $x^\prime$ has been used for training.}
    \label{fig:overview}
\end{figure*}

\subsection{Threat Model and Attack Taxonomy}
In this section, we give an overview of threat models we consider in this paper and describe a taxonomy of attacks leaking information from embedding models.
Figure~\ref{fig:overview} shows an overview of the attack taxonomy.


In many NLP applications, embedding vectors are often computed on sensitive user input.
Although existing frameworks propose computing embeddings locally on user's device to protect the privacy of the raw data~\cite{li2017privynet,osia2018deep,chi2018privacy,wang2018not}, the embeddings themselves are shared with machine learning service providers or other parties for downstream tasks (training or inference).
It is often tempting to assume that sharing embeddings might be ``safer'' than sharing the raw data, by their nature of being ``simply a vector of real numbers.'' However, this ignores the information that is retained in, and may be extracted from embeddings.
We investigate the following questions:~\textit{What kinds of sensitive information about the inputs are encoded in the embeddings? And can an adversary with access to the embeddings extract the encoded sensitive information?}

\paragraphb{Threat model.} 
Our threat model comprises the following entities. (1) $\datatrain$ is a training dataset that may contain sensitive information.
(2) $\enc$, the embedding model, which might be available in a white-box or black-box fashion. White-box access to the model reveals the model architecture and all parameters. Black-box access allows anyone to compute $\enc(x)$ for $x$ of their choice. (3) $\etarget = \{\enc(x_i^*)\}$, a set of embedding vectors on sensitive inputs $x_i^*$. (4) $\dataaux$ is an auxiliary dataset available to the adversary comprising of either limited labeled data drawn from the same distribution as $\datatrain$ or unlabeled raw text data.
In the text domain, unlabeled data is cheap to collect due to the enormous amount free text available on the Internet while labeling them is often much more expensive. 

An adversary, given access to some of the entities described above, aims to leak some sensitive information about the input $x$ from the embedding vector $\emb$. By looking at variants of what information an adversary possesses and their target, we arrive at three broad classes of attacks against embeddings.

With sensitive input data such as personal messages, it is natural to consider an adversary whose goal is to (partially) recover this text from $\emb$. Even if the input data is not sensitive, they may be associated with sensitive attributes, and an adversary could be tasked with learning these attributes from $\emb$. And finally, rather than recovering inputs, the adversary given some information about $\emb$ can aim to find if $x$ was used to train $\enc$ or not. These are formalized below.

\paragraphb{Embedding inversion attacks.} 
In this threat model, the adversary's goal is to invert a target embedding $\enc(x^*)$ and recover words in $x^*$. 
We consider attacks that involve both black-box and white-box access to $\enc$. 
The adversary is also allowed access to an unlabeled $\dataaux$ and in both scenarios is able to evaluate $\enc(x)$ for $x \in \dataaux$. 

\paragraphb{Sensitive attribute inference attacks.} In this threat model, the adversary's goal is to infer sensitive attribute $s^*$ of a secret input $x^*$ from a target embedding $\enc(x^*)$.
We assume the adversary has access to $\enc$, and a set of labeled data of the form $(x, s)$ for $x \in \dataaux$. 
We focus on discrete attributes $s^*\in\gS$, where $\gS$ is the set of all possible attribute classes, and an adversary performing the inference by learning a classifier $f$ on $\dataaux$.
Given sufficient labeled data , the adversary's task trivially reduces to plain supervised learning, which would rightly not be seen as adversarial. Therefore, the more interesting scenario, which is the focus of this paper, is when the adversary is given access to a very small set of labeled data (in the order of 10--50 per class) where transfer learning from $\emb$ is likely to outperform supervised learning directly from inputs $x$.

\paragraphb{Membership inference attacks.} Membership inference against ML models are well-studied attacks where the adversary has a target data point $x^*$ and the goal is to figure out with good accuracy whether $x^* \in \datatrain$ or not. 
Unlike previous attacks focused on supervised learning, some embedding models, such as word embeddings, allow you to trivially infer membership. For word embeddings, every member of a vocabulary set is necessarily  part of $\datatrain$.
Instead, we expand the definition of training data membership to consider this data with their \emph{contexts}, which is used in training. We assume the adversary has a target context of word $[w^*_1, \dots, w^*_n]$ and access to $\mV$ for word embedding, or a context of target sentences $(x^*_a, x^*_b)$ and access to the model $\enc$ for sentence embedding, and the goal is to decide the membership for the context.
We also consider the target to be an aggregated level of data sentences $[x_1^*, \ldots, x_n^*]$ comprising multiple contexts for the adversary to determine if it were part of training $\enc$.
Membership privacy for aggregation in the user level has also been explored in prior works~\cite{mcmahan2017learning,song2019auditing}.

We further assume adversary has a limited $\dataaux$ labeled with membership. 
We propose that this is a reasonable and practical assumption as training data for text embeddings are often collected from the Internet where an adversary can easily inject data or get access to small amounts of labeled data through a more expensive labeling process.
The assumption also holds for adversarial participants in collaborative training or federated learning~\cite{mcmahan2017communication,mcmahan2017learning}.

\section{Embedding Inversion Attacks} \label{sec:inversion}
The goal of inverting text embeddings is to recover the target  input texts $x^*$ from the embedding vectors $\enc(x^*)$ and access to $\enc$.
We focus on inverting embeddings of short texts and for practical reasons, it suffices to analyze the privacy of the inputs by considering attacks that recover a set of words without recovering the word ordering. We leave open the problem of recovering exact sequences and one promising direction involves language modeling~\cite{schmaltz2016word}.

A na\"ive approach for inversion would be enumerating all possible sequences from the vocabulary and find the recovery $\hat{x}$ such that $\enc(\hat{x}) = \embt$. 
Although such brute-force search only requires black-box access to $\enc$, the search space grows exponentially with the length of $x$ and thus inverting by enumerating is computationally infeasible.  

The brute-force approach does not capture inherently what information might be leaked by $\enc$ itself. This naturally raises the question of what attacks are possible if the adversary is given complete access to the parameters and architecture of $\enc$, i.e., white-box access. 
This also motivates a relaxation technique for optimization-based attacks in Section~\ref{sec:whitebox}.
We also consider the more constrained black-box access scenario in Section~\ref{sec:blackbox} where we develop learning based attacks that are much more efficient than exhaustive search by utilizing auxiliary data. 


\subsection{White-box Inversion}
\label{sec:whitebox}
\begin{algorithm}[t]
\caption{White-box inversion}
\begin{algorithmic}[1]
\State \textbf{Input:} target embedding $\embt$, white-box embedding model $\enc$ with lower layer representation function $\Psi$, temperature $\temp$, sparsity threshold $\threshsparse$, auxiliary dataset $\dataaux$
\State Query function $\enc$ and $\Psi$ with $\dataaux$ and collect $\{(\enc(x_i), \Psi(x_i)) | x_i\in\dataaux \}$.
\State Train a linear mapping $M$ by minimizing  $||M(\enc(x_i)) - \Psi(x_i) ||_2^2$ on $\{(\enc(x_i), \Psi(x_i))\}_i$.
\If{$\Psi$ is mean-pooling on word embedding $\mV$ of $\enc$}
\State Initialize $\vz\in\R^{|\gV|}$
\While{objective function of Eq~\ref{eq:opt3} not converged}
\State Update $\vz$ with gradient of Eq~\ref{eq:opt3}.
\State Project $\vz$ to non-negative orthant.
\EndWhile
\State\Return $\hat{x} = \{w_i | \vz_i \ge \threshsparse\}_{i=1}^{|\gV|}$ 
\Else
\State Initialize $\mZ=[\vz_1, \dots, \vz_\ell]\in\R^{\ell\times|\gV|}$
\While{objective function of Eq~\ref{eq:opt2} not converged}
\State Update $\mZ$ with gradient of Eq~\ref{eq:opt2}.
\EndWhile
\State\Return $\hat{x} = \{w_i | i = \argmax{\vz_j}\}_{j=1}^{\ell}$ 
\EndIf
\end{algorithmic}
\label{algo:whitebox}
\end{algorithm}

In a white-box scenario, we assume that the adversary has access to the embedding model $\enc$'s parameters and architecture. We formulate white-box inversion as the following optimization problem:
\begin{align}  \vspace{-0.1in}
\label{eq:opt0}
    \min_{\hat{x}\in \gX(\gV)} ||\enc(\hat{x}) - \embt ||_2^2
\end{align}
where $\gX(\gV)$ is the set of all possible sequences enumerated from the vocabulary $\gV$. The above optimization can be hard to solve directly due to the discrete input space. 
Inspired by prior work on relaxing categorical variables~\cite{jang2016categorical},
we propose a continuous relaxation of the sequential word input that allows more efficient optimization based on gradients. 

The goal of the discrete optimization in~\Eqref{eq:opt0} is to select a sequence of words such that the distance between the output embeddings is minimized. 
We relax the word selection at each position of the sequence with a continuous variable $\vz_i \in \sR^{|\vocab|}$.
As mentioned before, $\enc$ first maps the input $x$ of length $\ell$ into a sequence of word vectors $\mX = [\vv_1, \dots, \vv_\ell]$ and then computes the text embedding based on $\mX$. 
For optimizing $\vz_i$, we represent the selected word vectors $\hat{\vv}_i $ using a softmax attention mechanism:
\begin{align}
\label{eq:relax}
    \hat{\vv}_i = \mV^\top \cdot \softmax(\vz_i / \temp) \quad \textrm{for } i = 1, \dots, \ell
\end{align}
where $\mV$ is the word embedding matrix in $\enc$ and $\temp$ is a temperature parameter. The softmax function approximates hard argmax selection for $\temp < 1$. Intuitively, $\softmax(\vz_i / \temp)$ models the probabilities of selecting each word in the vocabulary at position $i$ in the sequence and $\hat{\vv}^i$ is the average of all word vectors weighted by the probabilities.

Let $\mZ = [\vz_1, \dots, \vz_\ell] \in\sR^{\ell \times |\gV|}$ and $\textrm{relaxed}(\mZ, \temp) = [\hat{\vv}_1, \dots, \hat{\vv}_\ell]$ be the sequence of softmax relaxed word vectors.
For simplicity, we denote $\enc(\mX)$ as the text embedding computed from any sequence of word vectors $\mX$. Then our relaxed optimization problem is: 
\begin{align}
\label{eq:opt1}
    \min_{\mZ}  ||\enc(\textrm{relaxed}(\mZ, \temp)) - \embt ||_2^2.
\end{align}
With continuous relaxation and white-box access to $\enc$, the optimization problem can now be solved by gradient-based methods as gradients of $\mZ$ can be calculated using back-propagation. To recover, we compute $\hat{x} = \{w_i | i = \argmax{\vz_j}\}_{j=1}^{\ell}$.

\paragraphb{Inverting embedding from deep models.} Prior works~\cite{mahendran2015understanding,ulyanov2018deep,dosovitskiy2016inverting,dosovitskiy2016generating} demonstrated that inverting image features from the higher layers of deep model can be challenging as representations from the higher layers are more abstract and generic while inverting from lower-layer representations results in better reconstruction of the image.  
With the recent advance in Transformer models~\cite{vaswani2017attention}, text embeddings can also be computed from deep models with many layers such as BERT~\cite{devlin2019bert}. 
Directly inverting such deep embeddings using the relaxed optimization method in~\Eqref{eq:opt1} can results in inaccurate recovery as the optimization becomes highly non-convex and different sequences with similar semantics can be mapped to the same location in the high-level embedding space. 
In reality, however, it is more common to use the embeddings from the higher layers than from lower layers for downstream tasks and thus an adversary might not be able to observe embeddings from lower layers at all. 

To resolve this issue, we propose to invert in two stages: (1) the adversary first maps the observed higher layer embedding $\emb$ to a lower layer one with a learned mapping function $M$, and (2) the adversary then solves the optimization problem of minimizing $||\Psi(\hat{x}) - M(\embt)||_2^2$ where $\Psi$ denotes the lower layer embedding function. 
To learn the mapping function $M$, adversary queries the white-box embedding model to get ($\enc(x_i), \Psi(x_i)$) for each auxiliary data $x_i\in\dataaux$ and trains $M$ with the set $\{(\enc(x_i), \Psi(x_i))\}_i$. 
After $M$ is trained, adversary solve the relaxed optimization:
\begin{align}
\label{eq:opt2}
    \min_{\mZ}  ||\Psi(\textrm{relaxed}(\mZ, \temp)) - M(\embt) ||_2^2
\end{align}
In practice, we find that learning a linear least square model as $M$ works reasonably well.

\paragraphb{A special case of inverting lowest representation.} 
The lowest embedding we can compute from the text embedding models would be the average of the word vectors, i.e., $\Psi(x) = (1/\ell) \cdot \sum_{i=1}^\ell \vv_i$. 
Inverting from such embedding reduces to the problem of recovering the exact words vectors with given averaged vector. 
Instead of using the relaxed optimization approach in~\Eqref{eq:opt1}, we use a sparse coding~\cite{olshausen1997sparse} formulation as following:
\begin{align}
\label{eq:opt3}
    &\min_{\vz\in\sR_{\ge 0}^{|\gV|}} || \mV^\top \cdot \vz - M(\embt) ||_2^2 + \lambda_\textrm{sp} ||\vz||_1
\end{align}
where $\mV$ is the word embedding matrix and the variable $\vz$ quantifies the contribution of each word vector to the average. 
We constrain $\vz$ to be non-negative as a word contributes either something positive if it is in the sequence $x$ or zero if it is not in $x$.
We further penalize $\vz$ with $\normlone$ norm to ensure its sparsity as only few words from the vocabulary contributed to the average.
The above optimization can be solved efficiently with projected gradient descent, where we set the coordinate $\vz_j$ to 0 if $\vz_j < 0$ after each descent step. 
The final recovered words are those with coefficient $\vz_j > \threshsparse$ for a sparsity threshold hyper-parameter $\threshsparse$. 

\subsection{Black-box Inversion}
\label{sec:blackbox}
In the black-box scenario, we assume that the adversary only has query access to the embedding model $\enc$, i.e., adversary observes the output embedding $\emb$ for a query $x$.
Gradient-based optimization is not applicable as the adversary does not have access to the model parameters and thus the gradients. 

Instead of searching for the most likely input, we directly extract the input information retained in the target embedding by formulating a learning problem.
The adversary learns an inversion model $\enci$ that takes a text embedding $\emb$ as input and outputs the set of words in the sequence $x$.
As mentioned before, our goal is to recover the set of words in the input independent of their word ordering.
We denote $\gW(x)$ as the set of words in the sequence $x$.
The adversary utilizes the auxiliary dataset $\dataaux$ and queries the black-box $\enc$ and obtain a collection of ($\emb, \gW(x)$) for each $x\in\dataaux$. 
The adversary then trains the inversion model $\enci$ to maximize $\log P_\enci(\gW(x)|\emb)$ on the collected set of ($\emb, \gW(x)$) values.
Once $\enci$ is trained, the adversary predicts the words in the target sequence $x^*$ as $\enci(\embt)$ for an observed $\embt$.

\paragraphb{Multi-label classification.}
The goal is to predict the set of words in sequence $x$ given the embedding $\emb$. 
A common choice for such a goal is to build a multi-label classification (MLC) model, where the model assigns a binary label of whether a word is in the set for each word in the vocabulary. 
The training objective function is then:
\begin{align}
\label{eq:mlc}
    \Ls_{\textrm{MLC}} 
    &= -\sum_{w\in\gV}[y_w\log(\hat{y}_w) + (1 - y_w)\log(1 - \hat{y}_w)]
\end{align}
where $\hat{y}_w = P_\enci(y_w|\emb)$ is the predicted probability of word $w$ given $\Upsilon$ conditioned on $\emb$ and $y_w = 1$ if word $w$ is in $x$ and 0 otherwise. 

\paragraphb{Multi-set prediction.}
One drawback in the above multi-label classification model is that the model predicts the appearance of each word independently. 
A more sophisticated formulation would be predicting the next word given the current predicted set of words until all words in the set are predicted. 
We adopt the multi-set prediction loss~\cite{welleck2018loss} (MSP) that is suited for the formulation. 
Our MSP model trains a recurrent neural network that predicts the next word in the set conditioned on the embedding $\emb$ and current predicted set of words. The training objective is as follows:
\begin{align}
\label{eq:msp}
\Ls_{\textrm{MSP}} = \sum_{i=1}^\ell\frac{1}{|\gW_i|}\sum_{w\in \gW_i}-\log P_\enci(w |\gW_{<i},\emb)
\end{align}
where $\gW_i$ is the set of words left to predict at timestamp $i$ and $\gW_{<i}$ is the set of the predicted words before $i$. 
The MSP formulation allows $\enci$ to learn a policy on the order of the words should be predicted instead of predicting all words independently and simultaneously.
In Section~\ref{sec:invesion_results}, we empirically show that MSP outperforms MLC in terms of the precision-recall trade-off.

\begin{algorithm}[t]
\caption{Black-box Inversion with multi-set prediction}
\begin{algorithmic}[1]
\State \textbf{Input:} target embedding $\embt$, black-box model $\enc$,  auxiliary data $\dataaux$
\Procedure{MSPLoss}{$x, \enc,\enci$}
\State Initialize $\Ls \gets 0, \gW_i \gets \gW(x), \gW_{<i}\gets \emptyset$
\For{$i=1$ to $\ell$}
\State Predict a word $\hat{w} = \argmax P_\enci(w | \gW_{<i}, \emb)$.
\State $\gW_i\gets\gW_i / \{\hat{w}\}$ and $\gW_{<i} \gets \gW_{<i}\cup\{\hat{w}\}$.
\State $\Ls\gets\Ls- \frac{1}{|\gW_i|}\sum_{w\in \gW_i}\log P_\enci(w |\gW_{<i},\emb)$.
\EndFor
\State\Return$\Ls$
\EndProcedure
\State Initialize $\enci$ as a recurrent neural network.
\While{$\enci$ not converged}
\State Sample a batch $\gB\subset\data_\textrm{aux}$.
\State Compute $\Ls_\textrm{MSP}\gets\frac{1}{|\gB|}\sum_{x_i\in\gB}\textproc{MSPLoss}(x_i, \enc,\enci)$.
\State Update $\enci$ with $\nabla\Ls_\textrm{MSP}$.
\EndWhile
\State\Return$\hat{x} = \{\argmax P_\enci(w | \gW{<i}, \embt)\}_{i=1}^\ell$
\end{algorithmic}
\label{algo:blackbox}
\end{algorithm}

\section{Attribute Inference Attacks}
\label{sec:attribute}
\begin{algorithm}[t]
\caption{Sensitive attribute inference}
\begin{algorithmic}[1]
\State \textbf{Input:} target embedding $\embt$, black-box model $\enc$, labeled auxiliary data $\dataaux$
\State Query $\enc$ with $\dataaux$ and collect $\{(\enc(x_i), s_i) | x_i\in\dataaux \}$.
\State Train a classifier $f$ that predicts $s$ on $\{(\enc(x_i), s_i) \}$.
\State\Return $\hat{s}=f(\embt)$
\end{algorithmic}
\end{algorithm}

Embeddings are designed to encode rich semantic information about the input data.
When the input data is user-related, e.g., a user's video watch history, the embedding vector naturally captures information about the user and is often much more informative than the raw input.    
Although in many applications such rich information from the embeddings is desired in order to provide personalized services to the user,
the embeddings may potentially reveal sensitive information about the input that might not directly appear in or be easy to infer from the input. As a motivating example, consider an adversary that may curate a small set of public comments labeled with authors of interest and then planning to use semantically rich embeddings on unlabeled, targeted text fragments to aim to deanonymize the author of the text. This is also the typical setup in stylometry research~\cite{ruder2016character,shetty2018a4nt} that we make more realistic by considering the challenges of curating labeled data (which may be a costly process) and strictly limiting the number of labeled examples the adversary has to work with.

For inferring sensitive attributes, we assume that the adversary has a limited set $\dataaux=\{(\enc(x_i), s_i)\}_i$ of embeddings labeled with the sensitive attribute, such as text authorship. 
The adversary then treats the inference problem as a downstream task and trains a classifier which predicts $s$ given $\emb$ as inputs on $\dataaux$.
At inference time, adversary simply applies the classifier on observed embedding $\enc(x^*)$ to infer the sensitive attribute of $x^*$.
We focus on the scenario of the adversary only having limited labeled data so as to
(a) closely match real scenarios where labeled sensitive data would be challenging to collect, and 
(b) demonstrate how easily sensitive information can be extracted from the embeddings which therefore constitutes an important vector of information leakage from embeddings.


\paragraphb{Connections between leakage and the objective.}
In \emph{supervised} learning, deep representations can reveal sensitive attributes of the input as these attributes might be used as internal features for the learning task~\cite{song2019overlearning}.
The connection between \emph{unsupervised} learning tasks and the leakage of sensitive attributes is less well understood.
The objective functions (Equation~\ref{eq:word} and~\ref{eq:sent}) that maximize the semantic similarity of data in context for training unsupervised dual-encoder embedding models 
fall into \emph{contrastive learning} framework~\cite{saunshi2019theoretical}.
This framework theoretically explains how unsupervised embedding training helps with the downstream tasks with a utility perspective. We explore how it might also favor the inference of some sensitive attributes from the perspective ofprivacy.

In this framework, training data of the embedding models are associated with latent classes (e.g., authors of the texts). 
When training with contrastive loss, the embeddings are learned so as to be similar for data in the same context and to be dissimilar for data coming from negative samples. Our approach takes advantage of the fact that data sharing the same latent class will often appear in the same context. Therefore, embedding similarity will be closer for inputs from  the same class, and consequently with the same sensitive attribute when there is a correlation. 
We further note that unsupervised embeddings are especially helpful for attribute inference under the limited data constraints as the embeddings are trained on much larger unlabeled data which allows them to learn semantic similarity over latent classes that might not be captured only given limited labeled data.
\section{Membership Inference Attacks} \label{sec:membership}
Both inverting embeddings and inferring sensitive attributes concern inference-time input privacy, i.e., information leaked about the input $x$ from the embedding vector. 
Another important aspect of privacy of ML models is \emph{training data privacy}, namely, what information about the training data (which might be potentially sensitive) is leaked by a model during the training process?
We focus on membership inference attacks~\cite{shokri2017membership} as a measurement of training data leakage in the embedding models.

The goal of membership inference is to infer whether a data point is in the training set of a given machine learning model. 
Classic membership inference attacks mainly target supervised machine learning, where a data point consists of a input feature vector and a class label. 
For unsupervised embedding models trained on units of data in context, we thus wish to infer the membership of a context of data (e.g., a sliding window of words or a pair of sentences). 

\subsection{Word Embeddings}
\begin{algorithm}[t]
\caption{MIA on word embeddings}
\begin{algorithmic}[1]
\State \textbf{Input:}  target window of words $\gC = [w_b, \dots, w_0, \dots, w_e]$, word embedding matrix $\mV$, similarity function $\delta$
\State Map words in $\gC$ with $\mV$ and get $[\vv_{w_b}, \dots, \vv_{w_0}, \dots, \vv_{w_e}].$
\State $\Delta\gets\{\delta_{i} | \delta_{i} = \delta(\vv_{w_0}, \vv_{w_i}), \forall w_i\in\gC / \{w_0\} \}$.
\State\Return ``member'' if $\frac{1}{|\Delta|}\sum_{\delta_{i}\in\Delta}\delta_{i} \ge \threshmia$ else ``non-member''
\end{algorithmic}
\end{algorithm}


Prior works~\cite{sablayrolles2019white,yeom2018privacy} on membership inference suggest that simple thresholding attacks based on loss values can be theoretically optimal under certain assumptions and practically competitive to more sophisticated attacks~\cite{shokri2017membership}.
In embedding models, the loss is approximated based on sampling during training as described in Section~\ref{sec:prelim} and computing exact loss is inefficient. 
We thus develop simple and efficient thresholding attacks based on similarity scores instead of loss values. 

Word embeddings are trained on a sliding window of words in the training corpus. 
To decide the membership for a window of words $\gC = [w_b, \dots, w_0, \dots, w_e]$, the adversary first converts each word into its embedding vectors $[\vv_{w_b}, \dots, \vv_{w_0}, \dots, \vv_{w_e}]$. 
Then the adversary computes a set of similarity scores $\Delta = \{\simscore(\vv_{w_0}, \vv_{w_i}) | \forall w_i\in\gC / \{w_0\}\}$, where $\simscore$ is a vector similarity measure function (e.g. cosine similarity).
Finally, the adversary uses the averaged score in $\Delta$ to decide membership: if the averaged score is above some threshold then $\gC$ is a member of the training data and not a member otherwise. 

\subsection{Sentence Embeddings}
\begin{algorithm}[t]
\caption{Aggregated-level MIA on sentence embeddings}
\begin{algorithmic}[1]
\State \textbf{Input:}  target sentences in context $\gX = [x_1, \dots, x_n]$, sentence embedding model $\enc$, similarity function $\delta$, auxiliary data $\dataaux$ with membership labels
\State Map sentences in $\gX$ with $\enc$ and get $[\enc(x_1), \dots, \enc(x_n)].$
\If{learning similarity function}
\State Initialize similarity function $\delta^\prime$ with projection $W_m$.
\While{$\delta^\prime$ not converged}
\State Sample a batch $\gB\subset\dataaux$.
\State Compute loss $\Ls_\textrm{MIA}$ for $(x_a, x_b)\in\gB$ with Eq~\ref{eq:mia}.
\State Update $W_m$ with $\nabla\Ls_\textrm{MIA}$.
\EndWhile
\State Replace similarity function $\delta\gets\delta^\prime$.
\EndIf
\State $\Delta\gets\{\delta_{i} | \delta_{i} = \delta(\enc(x_i), \enc(x_{i+1})) \}_{i=1}^{n-1}$.
\State\Return ``member'' if $\frac{1}{|\Delta|}\sum_{\delta_{i}\in\Delta}\delta_{i} \ge \threshmia$ else ``non-member''
\end{algorithmic}
\end{algorithm}

In sentence embeddings, we wish to decide membership of a pair of sentence in context ($x_a, x_b$). 
We simply use the similarity score $\simscore(\enc(x_a), \enc(x_b))$ as the decision score for membership inference as sentences in context used for training will be more similar to each other than sentences which were not used for training.

\paragraphb{Aggregate-level membership inference.} 
Sometimes, deciding membership of a pair of sentences might not be enough to cause a real privacy threat.
In many user-centric applications, models are trained on aggregation of data from users; e.g., a keyboard prediction model is trained on users' input logs on their phone~\cite{mcmahan2017learning}. 
In this scenario, the adversary infers membership on aggregate data from a particular user to learn whether this user participated training or not.

To perform MIA on aggregate text of $n$ sentences $\gX = [x_1, \allowbreak x_2, \allowbreak \dots, \allowbreak x_n]$, the adversary first gets each sentence embeddings $[\enc(x_1), \allowbreak  \enc(x_2), \allowbreak \dots, \allowbreak \enc(x_n)]$. Then the adversary collects the set of similarity score $\Delta = \{\simscore(x_i, x_{i + 1})\}_{i=1}^{n-1}$. 
Finally, similar to MIA against word embeddings, we use the average score in $\Delta$ as the decision score for membership inference.

\paragraphb{Learned similarity metric function.} Using a pre-defined similarity measure may not achieve best membership inference results.
With auxiliary data labeled with membership information, an adversary can learn a similarity metric customized for inference membership. More specifically, they learn a projection matrix $W_m$ and computes the learned similarity as $\simscore^\prime(x_a, x_b) = \simscore(W_m^\top\cdot\enc(x_a), W_m^\top\cdot\enc(x_b))$. The attack optimizes the binary cross-entropy loss with membership labels as following:
\begin{align}
\label{eq:mia}
    \Ls_\textrm{MIA} = -[y_m\log(\simscore^\prime_{a,b}) + (1-y_m) \log(1-\simscore^\prime_{a,b})] 
\end{align}
where $\simscore^\prime_{a,b}=\simscore^\prime(x_a, x_b)$ and $y_m = 1$ if $(x_a, x_b)$ where in the training set and 0 otherwise. 

\section{Experimental Evaluation} \label{sec:expt}
\subsection{Embedding Models and Datasets}
\label{sec:models}
As each of the attacks assess different perspectives of privacy, we evaluate them on different text embedding models that are either trained locally on consumer hardware or are trained elsewhere and public available.
Here, we describe text embedding models (and their corresponding datasets) we trained locally and evaluated against attacks described in previous sections. 
Other public models and datasets are detailed in subsequent subsections.

\paragraphb{Word Embeddings on Wikipedia.} We collected nearly 150,000 Wikipedia articles~\cite{mahoney2011large} for evaluating word embeddings.
We locally trained Word2Vec~\cite{mikolov2013distributed}, FastText~\cite{bojanowski2017enriching}, and GloVe~\cite{pennington2014glove} embedding models using half of the articles and use the other half for evaluating membership inference.

For all word embeddings, we set the number of dimension in embedding vector $d$ to be 100.
For Word2Vec and FastText, we set the number of sampled negative words $|\gV_\textrm{neg}|$ to be 25, learning rate to be 0.05, sliding window size to be 5 and number of training epochs to be 5.
For GloVe, we set the number of training iterations to be 50 as suggested in the original paper~\cite{pennington2014glove}. 

\paragraphb{Sentence Embeddings on BookCorpus.} Following prior works on training unsupervised sentence embeddings~\cite{kiros2015skip,logeswaran2018an}, we collected sentences from BookCorpus~\cite{zhu2015aligning} consists of 14,000 books.
We sample 40 millions sentences from half of the books as training data and use the rest as held-out data.

We locally train sentence embedding models with dual-encoder architecture described in Section~\ref{sec:prelim}. 
We considered two different neural network architecture for the embedding models: a recurrent neural network (LSTM~\cite{hochreiter1997long}) and a three-layer Transformer~\cite{vaswani2017attention}.
For the LSTM, we set the size of embedding dimension $d$ to be 1,200, number of training epochs to be 1 and learning rate to be 0.0005 following previous implementation~\cite{logeswaran2018an}.
For the Transformer, we set $d$ to be 600, number of training epochs to be 5 with a warm-up scheduled learning rate following~\cite{vaswani2017attention}. 
For both architectures, we train the model with Adam optimizer~\cite{kingma2014adam} and set the negative samples $\gX_\textrm{neg}$ as the other sentences in the same training batch at each step and $|\gX_\textrm{neg}| = 800$.

\subsection{Embedding Inversion}
\label{sec:invesion_results}
\paragraphb{Target and auxiliary data.} We randomly sample 100,000 sentences from 800 authors in the held-out BookCorpus data as the target data $\datatarget = \{x_i^*\}$ to be recovered and perform inversion on the set of embeddings $\etarget=\{\enc(x_i^*)\}$.
We consider two types of auxiliary data $\dataaux$: same-domain and cross-domain data. 
For same-domain $\dataaux$, we use a set of 200,000 randomly sampled sentences from BookCorpus that is disjoint to $\datatarget$.
For cross-domain $\dataaux$, we use a set of 800,000 randomly sampled sentences from Wikipedia articles. 
We set the number of cross-domain data points to be more than the same-domain data to match real-world constraints where cross-domain data is typically public and cheap to collect. 

\paragraphb{Additional embedding models.} 
In addition to the two dual-encoder embedding models with LSTM and Transformer, we further experiment with popular pre-trained language models for sentence embedding. 
We consider the original BERT~\cite{devlin2019bert} and the state-of-the-art ALBERT~\cite{lan2019albert}.  
We use mean pooling of the hidden token representations as the sentence embedding as described in Section~\ref{sec:prelim_emb}. 

\paragraphb{Evaluation metrics.}
As the goal of inversion is to recover the set of words in the sensitive inputs, we evaluate our inversion methods
based on precision (the percentage of recovered words in the target inputs), recall (the percentage of words in the target inputs are predicted) and F1 score which is the harmonic mean between precision and recall.

\paragraphb{White-box inversion setup.} 
We evaluate the white-box inversion with~\Eqref{eq:opt1} and~\Eqref{eq:opt3}. 
For inversion with~\Eqref{eq:opt1}, we set the temperature $\temp$ to be 0.05. 
For inversion with~\Eqref{eq:opt3}, we set the $L^1$ penalty coefficient $\lambda_\textrm{sp}$ to be 0.1 and the sparsity threshold $\threshsparse$ to be 0.01. 
For both methods, we use Adam optimizer~\cite{kingma2014adam} for gradient descent with learning rate set to 0.001. 
The hyper-parameters are tuned on a subset of the adversary's auxiliary data.

\begin{table}[t]
    \centering
    \caption{White-box inversion results on sentence embeddings. Pre denotes precision and Rec denotes recall. We leave the cross-domain results for Equation~\ref{eq:opt1} as blank as no learning on auxiliary data is needed. The best results are in bold.}
    \begin{tabular}{l|rrr|rrr}
    \toprule
    &  \multicolumn{3}{c|}{Same domain}  &  \multicolumn{3}{c}{Cross domain} \\
    \textbf{\Eqref{eq:opt1}} &  Pre & Rec & F1  &  Pre & Rec & F1 \\
    \midrule
    LSTM & 56.93 & 56.54 & 56.74 & - & - & - \\
    Transformer & 35.74 & 35.44 & 35.59 & - & - & - \\
    BERT & 0.84 & 0.89 & 0.87 & - & - & - \\
    ALBERT & 3.36 & 2.95 & 3.14 & - & - & - \\
    \midrule
     \textbf{\Eqref{eq:opt3}} &  Pre & Rec & F1  &  Pre & Rec & F1 \\
    \midrule
    LSTM & 63.68 & 56.69 & \textbf{59.98} & 57.98 & 48.05 & 52.55 \\
    Transformer & 65.32 & 60.39 & \textbf{62.76} & 59.97 & 54.45 & 57.08 \\
    BERT & 50.28 & 49.17 & \textbf{49.72} & 46.44 & 43.73 & 45.05 \\
    ALBERT & 70.91 & 55.49 & \textbf{62.26} & 68.45 & 53.18 & 59.86 \\
    \bottomrule
    \end{tabular}
    \label{tab:whitebox}
\end{table}

\begin{figure}[t]
\centering
\begin{tikzpicture}
\begin{axis}[
title=\textbf{BERT}, 
xlabel=Layer index of $\enc$, 
ylabel=Inversion performance,
xmin=0, xmax=12, ymin=0, ymax=100,
height=0.3\textwidth, width=0.5\textwidth, 
legend style = {at={(0.975, 0.85)}, legend columns=3, anchor=east},
xtick=data, grid=both, name=ax1]
\pgfplotstableread{plots/inv_bert_white.txt}\mydata;
\addplot[thick, mark=*, color=blue, dashed, mark options={solid}] table
[x expr=\thisrow{layer},y expr=\thisrow{Precision}] {\mydata};
\addlegendentry{Eq~\ref{eq:opt1} Pre};
\addplot[thick, mark=triangle*, color=red, dashed, mark options={solid}] table
[x expr=\thisrow{layer},y expr=\thisrow{Recall}] {\mydata};
\addlegendentry{Eq~\ref{eq:opt1} Rec};
\addplot[thick, mark=square*, color=orange, dashed, mark options={solid}] table
[x expr=\thisrow{layer},y expr=\thisrow{F1}] {\mydata};
\addlegendentry{Eq~\ref{eq:opt1} F1};
\pgfplotstableread{plots/inv_bert_white_map.txt}\mydata;
\addplot[thick, mark=o, color=teal] table
[x expr=\thisrow{layer},y expr=\thisrow{Precision}] {\mydata};
\addlegendentry{Eq~\ref{eq:opt3} Pre};
\addplot[thick, mark=triangle, color=purple] table
[x expr=\thisrow{layer},y expr=\thisrow{Recall}] {\mydata};
\addlegendentry{Eq~\ref{eq:opt3} Rec};
\addplot[thick, mark=square, color=brown] table
[x expr=\thisrow{layer},y expr=\thisrow{F1}] {\mydata};
\addlegendentry{Eq~\ref{eq:opt3} F1};
\end{axis}
\end{tikzpicture}
\\
\begin{tikzpicture}
\begin{axis}[
title=\textbf{ALBERT}, 
xlabel=Layer index of $\enc$, 
ylabel=Inversion performance,
xmin=0, xmax=12, ymin=0, ymax=100,
height=0.3\textwidth, width=0.5\textwidth, 
legend style = {at={(0.975, 0.35)}, legend columns=3, anchor=east},
xtick=data, grid=both, name=ax1]
\pgfplotstableread{plots/inv_albert_white.txt}\mydata;
\addplot[thick, mark=*, color=blue, dashed, mark options={solid}] table
[x expr=\thisrow{layer},y expr=\thisrow{Precision}] {\mydata};
\addlegendentry{Eq~\ref{eq:opt1} Pre};
\addplot[thick, mark=triangle*, color=red, dashed, mark options={solid}] table
[x expr=\thisrow{layer},y expr=\thisrow{Recall}] {\mydata};
\addlegendentry{Eq~\ref{eq:opt1} Rec};
\addplot[thick, mark=square*, color=orange, dashed, mark options={solid}] table
[x expr=\thisrow{layer},y expr=\thisrow{F1}] {\mydata};
\addlegendentry{Eq~\ref{eq:opt1} F1};
\pgfplotstableread{plots/inv_albert_white_map.txt}\mydata;
\addplot[thick, mark=o, color=teal] table
[x expr=\thisrow{layer},y expr=\thisrow{Precision}] {\mydata};
\addlegendentry{Eq~\ref{eq:opt3} Pre};
\addplot[thick, mark=triangle, color=purple] table
[x expr=\thisrow{layer},y expr=\thisrow{Recall}] {\mydata};
\addlegendentry{Eq~\ref{eq:opt3} Rec};
\addplot[thick, mark=square, color=brown] table
[x expr=\thisrow{layer},y expr=\thisrow{F1}] {\mydata};
\addlegendentry{Eq~\ref{eq:opt3} F1};
\end{axis}
\end{tikzpicture}
\caption{Performance of embedding inversion on sentence embedding from different layers of BERT and ALBERT.
Pre denotes precision and Rec denotes recall.
The x-axis is the layer index denoting which layer the embeddings are computed. Index 0 is the lowest (bottom) layer and 12 is the highest (top) layer.}
\label{fig:bert}
\end{figure}
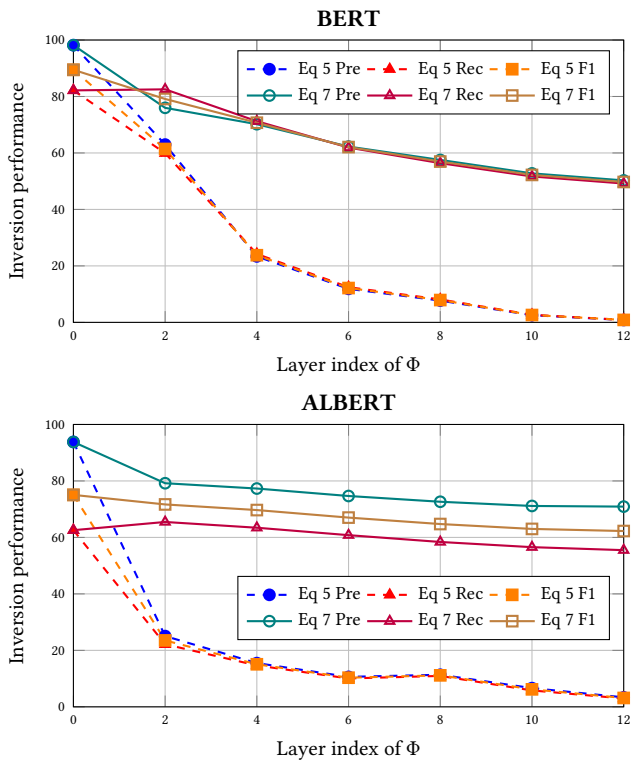

\paragraphb{White-box inversion results.} 
Table~\ref{tab:whitebox} summarizes the results. Note that there is no cross domain results for~\Eqref{eq:opt1} since no learning is needed. 
For inversion with~\Eqref{eq:opt1}, an adversary can extract more than a half and a third of the target input from LSTM and Transformer embedding models respectively as indicate by the F1 score.
This method performs poorly on BERT and ALBERT models. One plausible reason is that with many more layers in BERT where higher layer embeddings are more abstract than lower layers, directly optimizing for the highest layer could lead to recovering synonyms or semantically-related words rather than the targets.

For inversions with~\Eqref{eq:opt3}, all performance scores increase from~\Eqref{eq:opt1} on all models.
We can recover more than half of the input texts on nearly all models.
We also notice that there is only a little loss in performance when using cross-domain data for training the mapping $M$.

We further investigate the performance of inverting embeddings from different layers in BERT and ALBERT as shown in Figure~\ref{fig:bert}. 
There are in total 12 Transformer layers in BERT and ALBERT models and we choose embeddings from 2, 4, 6, 8, 10, 12 layer for inversion. 
We also compare with inverting from layer 0, i.e. mean-pooling of word embedding in BERT models.
The performance drops drastically when the layer goes high when inverting with~\Eqref{eq:opt1}. 
When training a mapping $M$ and inverting with~\Eqref{eq:opt3}, the drop in the performance is much less significant for higher layers.

\begin{table}[t]
    \centering
        \caption{Black-box inversion results on sentence embeddings. Pre denotes precision and Rec denotes recall. The best results are in bold.}
    \begin{tabular}{l|rrr|rrr}
    \toprule
     &  \multicolumn{3}{c|}{Same domain}  &  \multicolumn{3}{c}{Cross domain} \\
    \textbf{$\Ls_\textrm{MLC}$} &  Pre & Rec & F1  &  Pre & Rec & F1 \\
    \midrule
    LSTM & 90.53 & 39.35 & 54.86 & 87.71 & 32.91 & 47.86 \\
    Transformer & 81.18 & 26.07 & 39.47 & 77.34 & 21.82 & 34.04 \\
    BERT & 89.70 & 36.80 & 52.19 & 84.05 & 30.28 & 44.52 \\
    ALBERT & 95.92 & 48.71 & 64.61 & 92.51 & 44.30 & 59.91 \\
    \midrule
    \textbf{$\Ls_\textrm{MSP}$} &  Pre & Rec & F1  &  Pre & Rec & F1 \\
    \midrule
    LSTM & 61.69 & 64.40 & \textbf{63.02} & 59.52 & 62.20 & 60.83 \\
    Transformer & 53.59 & 55.72 & \textbf{54.63} & 51.37 & 52.78 & 52.07 \\
    BERT & 60.21 & 59.31 & \textbf{59.76} & 55.18 & 55.44 & 55.31 \\
    ALBERT & 76.77 & 72.05 & \textbf{74.33} & 74.07 & 70.66 & 72.32 \\
    \bottomrule
    \end{tabular}
    \label{tab:blackbox}
\end{table}

\paragraphb{Black-box setup.} 
We evaluate the black-box inversion with $\Ls_\textrm{MLC}$ (\Eqref{eq:mlc}) and $\Ls_\textrm{MSP}$ (\Eqref{eq:msp}).
For $\Ls_\textrm{MLC}$, we train $|\gV|$ binary classifiers as $\enci$ for each $w\in\gV$.
For $\Ls_\textrm{MSP}$, we train a one-layer LSTM as $\enci$ with number of hidden units set to 300. 
We train both models for 30 epochs with Adam optimizer and set learning rate to 0.001, batch size to 256. 

\paragraphb{Black-box results.} 
Table~\ref{tab:blackbox} summarizes the results. 
Inversion with $\Ls_\textrm{MLC}$ can achieve high precision with low recall. 
This might be due to the inversion model $\enci$ being biased towards the auxiliary data and thus confident in predicting some words while not others.
Inversion with $\Ls_\textrm{MSP}$ yields better balance 
precision and recall and thus higher F1 scores.

\subsection{Sensitive Attribute Inference}

\paragraphb{Target and auxiliary data.} We consider authorship of sentence to be the sensitive attribute and target data to be a collection of sentences of randomly sampled author set $\gS$ from the held-out dataset of BookCorpus, with 250 sentences per author.
The goal is to classify authorship $s$ of sentences amongst $\gS$ given sentence embeddings. For auxiliary data, we consider 10, 20, 30, 40 and 50 labeled sentences (disjoint from those in the target dataset) per author.
We also vary the size of author set $|\gS|=100, 200, 400$ and 800 where the inference task becomes harder as $|\gS|$ increases.

\paragraphb{Baseline model.} To demonstrate sensitive attribute leakage from the embedding vector, we compare the attack performance between embedding models and a baseline model that is trained from raw sentences without access to the embeddings. 
We train a TextCNN model~\cite{kim2014convolutional} as the baseline, which is efficient to train and has been shown to achieve accurate authorship attribution in previous works~\cite{ruder2016character,shetty2018a4nt}.

\paragraphb{Additional embedding models.} As discussed in Section~\ref{sec:attribute}, the embedding models trained with dual-encoder and contrastive learning that is a focus of this paper might, in particular, favor attribute inference.
We compare the dual-encoder embedding models with two other embedding models trained with different objective functions. 
The first is the Skip-thought embedding model~\cite{kiros2015skip} which is trained to generate the context given a sentence. 
We also evaluate on InferSent embeddings~\cite{conneau2017supervised} that is trained with supervised natural language inference tasks.

\paragraphb{Setup.}
For the baseline TextCNN model, we set the number of  filters in convolutional layer to 128.
For all other embedding models, we train a linear classifier for authorship inference.
We train all inference models for 30 epochs with Adam optimizer and set learning rate to 0.001 and batch size to 128. 
We repeat each experiment 5 times with sampled $\gS$ using different random seed and report the averaged top-5 accuracy.

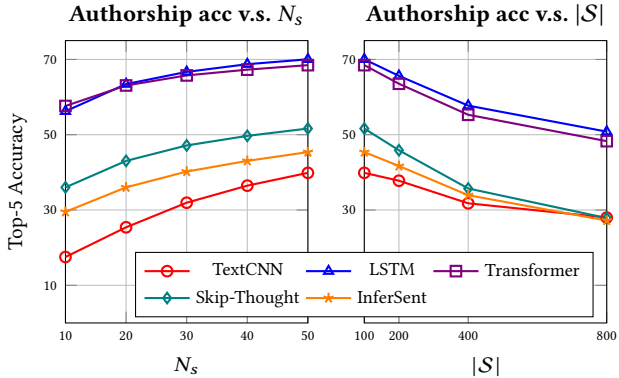
\begin{figure}[t]
\centering
\begin{tikzpicture}
\begin{axis}[
title=\textbf{Authorship acc v.s. $N_s$},
xlabel= $N_s$, 
ylabel=Top-5 Accuracy,
ytick={10,30,50,70},
xmin=10, xmax=50, ymin=0, ymax=75,
height=0.3\textwidth, width=0.27\textwidth, 
xtick=data, grid=both, name=ax1]
\pgfplotstableread{plots/attr.txt}\mydata;
\addplot[thick, mark=o, color=red] table
[x expr=\thisrow{ns},y expr=\thisrow{cnn}] {\mydata};
\addplot[thick, mark=triangle, color=blue] table
[x expr=\thisrow{ns},y expr=\thisrow{quick}] {\mydata};
\addplot[thick, mark=square, color=violet] table
[x expr=\thisrow{ns},y expr=\thisrow{trans}] {\mydata};
\addplot[thick, mark=diamond, color=teal] table
[x expr=\thisrow{ns},y expr=\thisrow{skip}] {\mydata};
\addplot[thick, mark=star, color=orange] table
[x expr=\thisrow{ns},y expr=\thisrow{infer}] {\mydata};
\end{axis}
\begin{axis}[
title=\textbf{Authorship acc v.s. $|\gS|$},
xlabel= $|\gS|$, 
xmin=100, xmax=800, ymin=0, ymax=75,
height=0.3\textwidth, width=0.27\textwidth, 
legend style={at={(0.,0.02)}, legend columns=3, anchor=south},
ytick={10,30,50,70},
xtick=data, grid=both,
name=ax2, at={(ax1.south east)}, xshift=0.75cm]
\pgfplotstableread{plots/attr_num_author.txt}\mydata;
\addplot[thick, mark=o, color=red] table
[x expr=\thisrow{ns},y expr=\thisrow{cnn}] {\mydata};
\addlegendentry{TextCNN};
\addplot[thick, mark=triangle, color=blue] table
[x expr=\thisrow{ns},y expr=\thisrow{quick}] {\mydata};
\addlegendentry{LSTM};
\addplot[thick, mark=square, color=violet] table
[x expr=\thisrow{ns},y expr=\thisrow{trans}] {\mydata};
\addlegendentry{Transformer};
\addplot[thick, mark=diamond, color=teal] table
[x expr=\thisrow{ns},y expr=\thisrow{skip}] {\mydata};
\addlegendentry{Skip-Thought};
\addplot[thick, mark=star, color=orange] table
[x expr=\thisrow{ns},y expr=\thisrow{infer}] {\mydata};
\addlegendentry{InferSent};
\end{axis}
\end{tikzpicture}
\caption{Performance of sensitive attribute (author) inference with different models. 
For the left figure, the x-axis is the number of labeled data per author $N_s$ and the y-axis is the top-5 accuracy of classifying 100 authors.
For the right figure, the x-axis is the number of author classes $|\gS|$ and the y-axis is the top-5 accuracy with 50 labeled data per author.}
\label{fig:author}
\end{figure}

\paragraphb{Results.} Figure~\ref{fig:author} demonstrates the results of authorship inference with different number of labeled data and different number of authors. 
The baseline TextCNN models trained from scratch with limited labeled data have the worst performance across all settings. 
Skip-thought and InferSent embeddings can outperform TextCNN but the gap between the performance decreases as the number of labeled examples increase. 
The LSTM and Transformer dual-encoder models achieve best inference results and are better than the baseline by a significant margin in all scenarios. 
This demonstrates that embeddings from dual-encoder models trained with contrastive learning framework~\cite{saunshi2019theoretical} aid attribute inference attacks the most comparing to other pre-trained embeddings.

\subsection{Membership Inference}
\paragraphb{Evaluation metrics.} We consider membership inference as a binary classification task of distinguishing members and non-members of training data. 
We evaluate the performance of membership inference attacks with adversarial advantage~\cite{yeom2018privacy}, defined as the difference between the true and false positive rate. 
Random guesses offer an advantage of 0.

\paragraphb{Word embedding setup.} We evaluate membership inference attacks on sliding window of 5 words from Wikipedia articles. 
We also perform the attack separately for windows with central words having different frequencies, following the intuition that rare words are prone to more memorization~\cite{song2019auditing}. 
Specifically, we evaluate the attack for windows with frequency of central words in decile (10th, $\dots$, 90th percentile) ranges. We use cosine similarity for $\pi$.

\begin{figure}[t]
\centering
\begin{tikzpicture}
\begin{axis}[
title=\textbf{MIA on word embedding}, 
xlabel=Inverse frequency percentile, 
ylabel=Adversarial advantage,
xmin=10, xmax=90, 
height=0.3\textwidth, width=0.5\textwidth, 
legend pos=north west,
xtick=data, grid=both]
\pgfplotstableread{plots/mia_word.txt}\mydata;
\addplot[thick, mark=o, color=blue] table
[x expr=\thisrow{freq},y expr=\thisrow{Word2Vec}] {\mydata};
\addlegendentry{Word2Vec};
\addplot[thick, mark=triangle, color=red] table
[x expr=\thisrow{freq},y expr=\thisrow{FastText}] {\mydata};
\addlegendentry{FastText};
\addplot[thick, mark=square, color=orange] table
[x expr=\thisrow{freq},y expr=\thisrow{GloVe}] {\mydata};
\addlegendentry{GloVe};
\end{axis}
\end{tikzpicture} \vspace{-0.2in}
\caption{ Performance of membership inference attack on Word2Vec, FastText and GloVe. The x-axis is the inverse frequency percentile range (the smaller the more frequent) and the y-axis is the adversarial advantage.}
\label{fig:mia_word}
\end{figure}
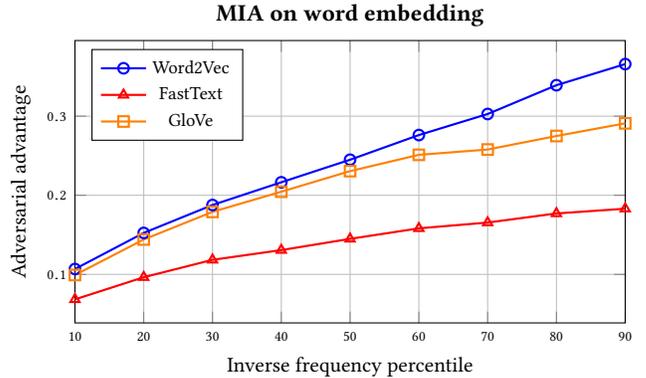

\paragraphb{Word embedding results.}
Figure~\ref{fig:mia_word} demonstrates the results.
For frequent (10th percentile) central words, there is almost no memorization (advantage $<0.1$). As this  frequency decreases, the advantage increases correspondingly to roughly $0.3$. FastText is most resistant to these attacks possibly due to its operating on sub-word units rather than exact words. 

\paragraphb{Sentence embedding setup.}
We evaluate membership inference attacks on context- and aggregate-level data from the BookCorpus dataset. We consider a pair of sentences for context-level data, and a collection of sentences from the same book for aggregate-level data with a goal of inferring if the book is part of the training corpus. As with word embeddings, we evaluate the attack on different frequencies (averaged across words in a sentence). We use dot-product similarity for $\pi$,  and for learning-based similarity, we learn a projection matrix with 10\% of training and hold-out data. We optimize  $\Ls_\textrm{MIA}$ with the Adam optimizer for 10 epochs with learning rate set to 0.001.

\begin{figure}[t]
\centering
\begin{tikzpicture}
\begin{axis}[
title=\textbf{LSTM}, 
xlabel=Inverse frequency percentile, 
ylabel=Adversarial advantage,
xmin=50, xmax=90,  ymax=0.375,
height=0.3\textwidth, width=0.27\textwidth, 
legend pos=north west,
legend columns=2,
xtick=data, grid=both, name=ax1]
\pgfplotstableread{plots/mia_lstm.txt}\mydata;
\addplot[thick, mark=o, color=blue] table
[x expr=\thisrow{freq},y expr=\thisrow{ctx}] {\mydata};
\addlegendentry{ctx};
\addplot[thick, mark=triangle, color=red] table
[x expr=\thisrow{freq},y expr=\thisrow{book}] {\mydata};
\addlegendentry{book};
\addplot[thick, mark=square, color=orange] table
[x expr=\thisrow{freq},y expr=\thisrow{lctx}] {\mydata};
\addlegendentry{ctx$^\prime$};
\addplot[thick, mark=diamond, color=teal] table
[x expr=\thisrow{freq},y expr=\thisrow{lbook}] {\mydata};
\addlegendentry{book$^\prime$};
\end{axis}
\begin{axis}[
title=\textbf{Transformer}, 
xlabel=Inverse frequency percentile, 
xmin=50, xmax=90,  ymax=0.375,
height=0.3\textwidth, width=0.27\textwidth, 
legend pos=north west,
xtick=data, grid=both,
name=ax2, at={(ax1.south east)}, xshift=0.75cm]
\pgfplotstableread{plots/mia_trans.txt}\mydata;
\addplot[thick, mark=o, color=blue] table
[x expr=\thisrow{freq},y expr=\thisrow{ctx}] {\mydata};
\addplot[thick, mark=triangle, color=red] table
[x expr=\thisrow{freq},y expr=\thisrow{book}] {\mydata};
\addplot[thick, mark=square, color=orange] table
[x expr=\thisrow{freq},y expr=\thisrow{lctx}] {\mydata};
\addplot[thick, mark=diamond, color=teal] table
[x expr=\thisrow{freq},y expr=\thisrow{lbook}] {\mydata};
\end{axis}
\end{tikzpicture} 
\caption{Performance of context-level (ctx) and book-level membership inference attack on sentence embedding trained with LSTM and Transformer.
The x-axis is the inverse frequency percentile range (the smaller the more frequent) and the y-axis is the adversarial advantage.
The ctx$^\prime$ and book$^\prime$ denote results with learned similarity}
\label{fig:mia_sent}
\end{figure}
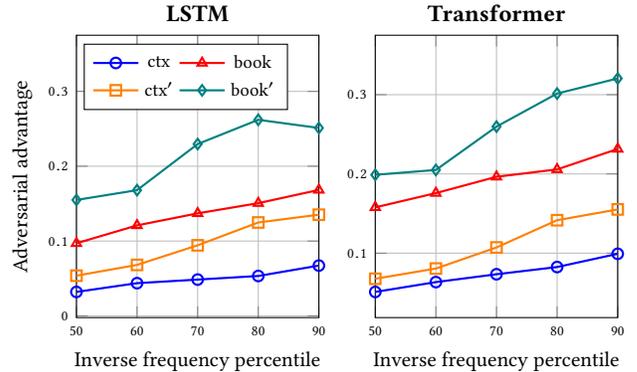

\paragraphb{Sentence embedding results.}
Figure~\ref{fig:mia_sent} shows the results of MIA on embeddings from LSTM and Transformer dual-encoder models.
For both models, context-level MIA advantage scores are below 0.1 for all frequency ranges, indicating that adversarial does not gain much information about context-level membership from the embeddings. 
Learning based similarity can improve context-level MIA slightly.
For aggregated book-level inference, adversaries achieve a greater advantage than context-level inference and learning-based similarity scores can boost the advantage to 0.3 for books with infrequent sentences.  
\section{Defenses}
\label{sec:defence}

\begin{figure}[t]
\centering
\begin{tikzpicture}
\begin{axis}[
title=\textbf{LSTM}, 
xlabel=$\lambda_w$, 
ylabel=Utility Score,
xmin=0, xmax=0.4, ymin=0.7,
height=0.3\textwidth, width=0.27\textwidth, 
legend pos=south west,
xtick=data, grid=both, name=ax1]
\pgfplotstableread{plots/util_word_lstm.txt}\mydata;
\addplot[thick, mark=o, color=blue] table
[x expr=\thisrow{lambda},y expr=\thisrow{MPQA}] {\mydata};
\addlegendentry{MPQA};
\addplot[thick, mark=triangle, color=red] table
[x expr=\thisrow{lambda},y expr=\thisrow{TREC}] {\mydata};
\addlegendentry{TREC};
\addplot[thick, mark=square, color=orange] table
[x expr=\thisrow{lambda},y expr=\thisrow{SUBJ}] {\mydata};
\addlegendentry{SUBJ};
\addplot[thick, mark=diamond, color=teal] table
[x expr=\thisrow{lambda},y expr=\thisrow{MSRP}] {\mydata};
\addlegendentry{MSRP};
\end{axis}
\begin{axis}[
title=\textbf{Transformer}, 
xlabel=$\lambda_w$, 
xmin=0, xmax=0.1,  ymin=0.7,
height=0.3\textwidth, width=0.27\textwidth, 
legend pos=south west, legend columns=2,
xtick=data, grid=both,
name=ax2, at={(ax1.south east)}, xshift=0.75cm]
\pgfplotstableread{plots/util_word_trans.txt}\mydata;
\addplot[thick, mark=o, color=blue] table
[x expr=\thisrow{lambda},y expr=\thisrow{MPQA}] {\mydata};
\addplot[thick, mark=triangle, color=red] table
[x expr=\thisrow{lambda},y expr=\thisrow{TREC}] {\mydata};
\addplot[thick, mark=square, color=orange] table
[x expr=\thisrow{lambda},y expr=\thisrow{SUBJ}] {\mydata};
\addplot[thick, mark=diamond, color=teal] table
[x expr=\thisrow{lambda},y expr=\thisrow{MSRP}] {\mydata};
\end{axis}
\end{tikzpicture}
\\
\begin{tikzpicture}
\begin{axis}[
title=\textbf{LSTM}, 
xlabel=$\lambda_w$, 
ylabel=Inversion F1 Score,
xmin=0, xmax=0.4,
height=0.3\textwidth, width=0.27\textwidth, 
legend pos=south west,
xtick=data, grid=both, name=ax1]
\pgfplotstableread{plots/util_word_lstm.txt}\mydata;
\addplot[thick, mark=o, color=blue] table
[x expr=\thisrow{lambda},y expr=\thisrow{White-box}] {\mydata};
\addlegendentry{White-box};
\addplot[thick, mark=triangle, color=red] table
[x expr=\thisrow{lambda},y expr=\thisrow{Multi-label}] {\mydata};
\addlegendentry{$\Ls_\textrm{MLC}$};
\addplot[thick, mark=square, color=orange] table
[x expr=\thisrow{lambda},y expr=\thisrow{Set-predict}] {\mydata};
\addlegendentry{$\Ls_\textrm{MSP}$};
\end{axis}
\begin{axis}[
title=\textbf{Transformer}, 
xlabel=$\lambda_w$, 
xmin=0, xmax=0.1, 
height=0.3\textwidth, width=0.27\textwidth, 
legend pos=south west, 
xtick=data, grid=both,
name=ax2, at={(ax1.south east)}, xshift=0.75cm]
\pgfplotstableread{plots/util_word_trans.txt}\mydata;
\addplot[thick, mark=o, color=blue] table
[x expr=\thisrow{lambda},y expr=\thisrow{White-box}] {\mydata};
\addplot[thick, mark=triangle, color=red] table
[x expr=\thisrow{lambda},y expr=\thisrow{Multi-label}] {\mydata};
\addplot[thick, mark=square, color=orange] table
[x expr=\thisrow{lambda},y expr=\thisrow{Set-predict}] {\mydata};
\end{axis}
\end{tikzpicture}
\caption{Effects of adversarial training against embedding inversion on the utility (top row) and the inversion F1 score (bottom row) for sentence embeddings trained with LSTM and Transformer.}
\label{fig:adv_inversion}
\end{figure}
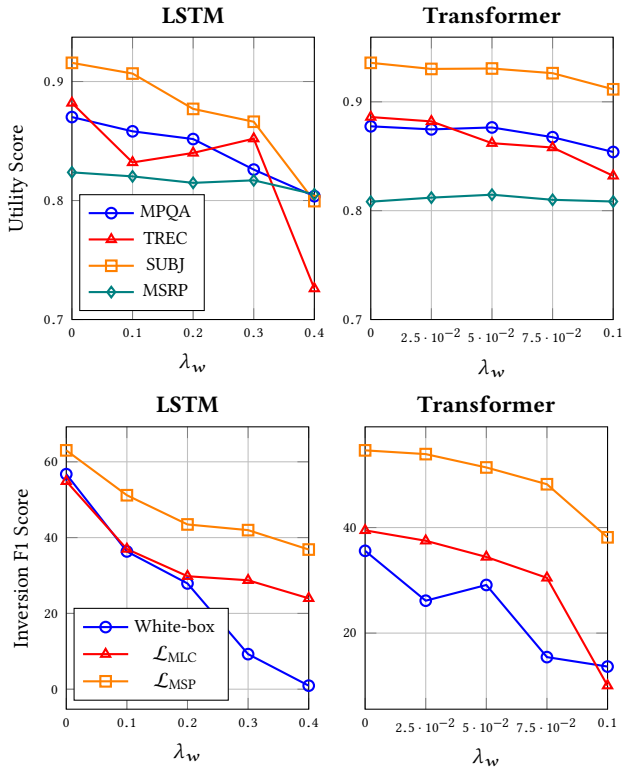

\begin{figure}[t]
\centering
\begin{tikzpicture}
\begin{axis}[
title=\textbf{LSTM}, 
xlabel=$\lambda_s$, 
ylabel=Utility Score,
xmin=0, xmax=1, ymin=0.5,
height=0.3\textwidth, width=0.27\textwidth, 
legend pos=south west,
xtick=data, grid=both, name=ax1]
\pgfplotstableread{plots/util_author_lstm.txt}\mydata;
\addplot[thick, mark=o, color=blue] table
[x expr=\thisrow{lambda},y expr=\thisrow{MPQA}] {\mydata};
\addlegendentry{MPQA};
\addplot[thick, mark=triangle, color=red] table
[x expr=\thisrow{lambda},y expr=\thisrow{TREC}] {\mydata};
\addlegendentry{TREC};
\addplot[thick, mark=square, color=orange] table
[x expr=\thisrow{lambda},y expr=\thisrow{SUBJ}] {\mydata};
\addlegendentry{SUBJ};
\addplot[thick, mark=diamond, color=teal] table
[x expr=\thisrow{lambda},y expr=\thisrow{MSRP}] {\mydata};
\addlegendentry{MSRP};
\end{axis}
\begin{axis}[
title=\textbf{Transformer}, 
xlabel=$\lambda_s$, 
xmin=0, xmax=0.6,  ymin=0.5,
height=0.3\textwidth, width=0.27\textwidth, 
legend pos=south west,
xtick=data, grid=both,
name=ax2, at={(ax1.south east)}, xshift=0.75cm]
\pgfplotstableread{plots/util_author_trans.txt}\mydata;
\addplot[thick, mark=o, color=blue] table
[x expr=\thisrow{lambda},y expr=\thisrow{MPQA}] {\mydata};
\addplot[thick, mark=triangle, color=red] table
[x expr=\thisrow{lambda},y expr=\thisrow{TREC}] {\mydata};
\addplot[thick, mark=square, color=orange] table
[x expr=\thisrow{lambda},y expr=\thisrow{SUBJ}] {\mydata};
\addplot[thick, mark=diamond, color=teal] table
[x expr=\thisrow{lambda},y expr=\thisrow{MSRP}] {\mydata};
\end{axis}
\end{tikzpicture}
\\
\begin{tikzpicture}
\begin{axis}[
title=\textbf{LSTM}, 
xlabel=$\lambda_s$, 
ylabel=Top-5 Accuracy,
xmin=0, xmax=1,
height=0.3\textwidth, width=0.27\textwidth, 
legend pos=south west,
xtick=data, grid=both, name=ax1]
\pgfplotstableread{plots/util_author_lstm.txt}\mydata;
\addplot[thick, mark=o, color=blue] table
[x expr=\thisrow{lambda},y expr=\thisrow{n50}] {\mydata};
\addlegendentry{$N_s$ = 50};
\addplot[thick, mark=triangle, color=red] table
[x expr=\thisrow{lambda},y expr=\thisrow{n10}] {\mydata};
\addlegendentry{$N_s$ = 10};
\end{axis}
\begin{axis}[
title=\textbf{Transformer}, 
xlabel=$\lambda_s$, 
xmin=0, xmax=0.6,
height=0.3\textwidth, width=0.27\textwidth, 
legend pos=south west, 
xtick=data, grid=both,
name=ax2, at={(ax1.south east)}, xshift=0.75cm]
\pgfplotstableread{plots/util_author_trans.txt}\mydata;
\addplot[thick, mark=o, color=blue] table
[x expr=\thisrow{lambda},y expr=\thisrow{n50}] {\mydata};
\addplot[thick, mark=triangle, color=red] table
[x expr=\thisrow{lambda},y expr=\thisrow{n10}] {\mydata};
\end{axis}
\end{tikzpicture} \vspace{-0.1in}
\caption{Effects of adversarial training against sensitive attribute inference on the utility (top row) and the author classification top-5 accuracy (bottom row) for sentence embeddings trained with LSTM and Transformer. $N_s$ denotes number of labeled data per author.}
\label{fig:adv_sensitive}
\end{figure}
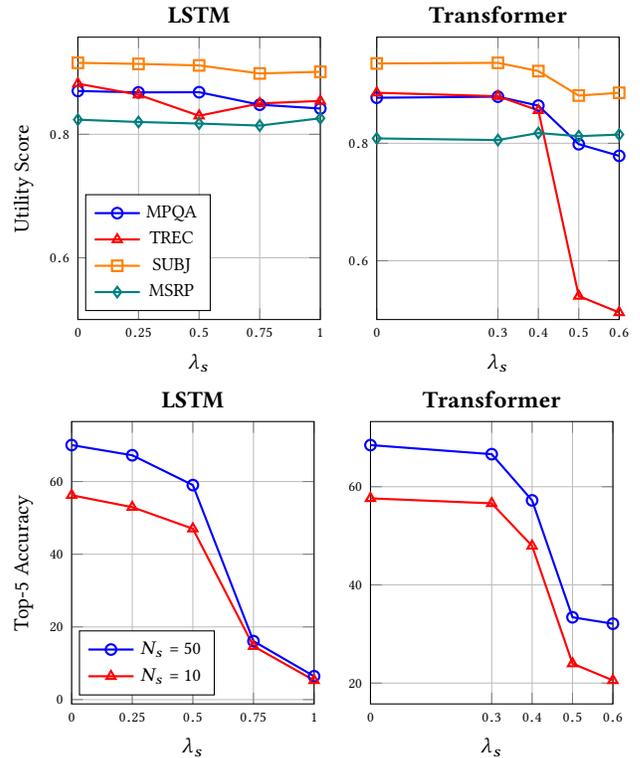

\paragraphb{Adversarial training.} Attacks involving embedding inversions and sensitive attributes are both inference-time attacks that wish to infer information about the sensitive inputs given the output of the embedding.
A common defence mechanism for such inference-time attacks is adversarial training~\cite{edwards2015censoring,xie2017controllable,li2018towards,coavoux2018privacy,elazar2018adversarial}. 
In this framework, a simulated adversary $\gA$ is trained to infer any sensitive information jointly with a main model $\enc$ while $\enc$ is trained to maximize the adversary's loss and minimize the primary learning objective.
The embeddings trained with this minimax optimization protects sensitive information from an inference-time adversary to an extent while maintaining their utility for downstream tasks.

To defend against embedding inversion attacks, $\gA$ is trained to predict the words in $x$ given $\emb$. 
For a pair of sentences in context ($x_a, x_b$) and a set of negative example $\gX_\textrm{neg}$, the training objective for $\enc$ is: \begin{align*}
\min_{\enc}\max_{\gA}  \lambda_w\log P_\gA(\gW(x_b) | \enc(x_b)) -\log P_\enc(x_b|x_a, \gX_\textrm{neg}),\end{align*} 
where $\gW(x)$ is the set of words in $x$ and the coefficient $\lambda_w$ controls the balance between the two terms. 
A natural choice for $\log P_\gA$ is to use the multi-label classification loss $\Ls_\textrm{MLC}$ in \Eqref{eq:mlc}. To defend against sensitive attribute attacks, $\gA$ is trained to predict the sensitive attribute $s$ in $x$ from the embedding $\emb$. As above, the training objective for $\enc$ is:
\begin{align*}
\min_{\enc}\max_{\gA}  \lambda_s\log P_\gA(s | \enc(x_b)) -\log P_\enc(x_b|x_a, \gX_\textrm{neg}), 
\end{align*}
where the coefficient $\lambda_s$ controls the balance between the two terms. 
Both minimax training objectives can be efficiently optimized with the gradient reversal trick~\cite{ganin2015unsupervised}. 

\paragraphb{Results with adversarial training.}
We evaluate this adversarial training approach on dual-encoder models with LSTM and Transformer. 
We keep all the training hyper-parameters the same as in Section~\ref{sec:models}.
We train multiple models under different $\lambda_w$ and $\lambda_s$ and evaluate their effects on attack performance as well as utility scores for downstream tasks.
For utility measurement, we evaluate the adversarially trained embeddings on four sentence analysis benchmarks: multi-perspective question answering (MPQA)~\cite{wiebe2005annotating}, text retrieval (TREC)~\cite{li2002learning}, subjectivity analysis (SUBJ)~\cite{pang2004sentimental} and Microsoft Research paraphrase corpus (MSRP)~\cite{dolan2005automatically}.
We treat all benchmarks as classification problem and train a logistic regression model for each task following previous works~\cite{kiros2015skip,logeswaran2018an}.

Figure~\ref{fig:adv_inversion} shows the results for adversarial training against inversion.
As $\lambda_w$ increases, the adversary performance drops for all models. White-box inversion attacks drop most significantly.
LSTM embedding models needs a larger $\lambda_w$ than Transformer models to achieve similar mitigation, possible due to the fact that Transformer models have more capacity and are more capable of learning two tasks.
Utility scores on the benchmarks drop more drastically on embedding models using LSTM than Transformer due to the larger value of $\lambda_w$.

Figure~\ref{fig:adv_sensitive} shows the results for adversarial training against authorship inference.
As $\lambda_s$ increases, the adversary performance on inferring authorship drops significantly for both inference attack models trained with 10 and 50 labeled data per author.
Nearly all utility scores on the four benchmarks remain rather stable for different $\lambda_s$'s. 
This also demonstrates that different adversary tasks can have different impact on the utility of embeddings.
In our case, removing input word information from embeddings is a harder task than removing authorship and thus will have a larger impact on the utility. 

\section{Related Work}
\paragraphb{Privacy in deep representations.}
Prior works demonstrate that representations from supervised learning models leak sensitive attributes about input data that are statistically uncorrelated with the learning task~\cite{song2019overlearning}, and gradient updates in collaborative training (which depend on hidden representations of training inputs) also leak sensitive attributes~\cite{melis2019exploiting}.
In contrast, this work focuses on leakage in unsupervised text embedding models and considers leakage of both sensitive attributes and raw input text.
In addition, we consider a more realistic scenario where the labeled data is limited for measuring sensitive attributes leakage which is not evaluated in prior works.

Recently and concurrently with this work, Pan et al.~\cite{pan2020privacy} also considered the privacy risks of general purpose language models and analyzed model inversion attacks on these models. Our work presents and develops a taxonomy of attacks, which is broader in scope than their work. Our work on inversion also assumes no structures or patterns in input text, which is the main focus of their work, and our work shows that we still recover substantial portion of input data. Additional structural assumptions will lead to a higher recovery rate as is shown in their work.  

There is a large body of research on learning privacy-preserving deep representations in supervised models. 
One popular approach is through adversarial training~\cite{edwards2015censoring, xie2017controllable} as detailed in Section~\ref{sec:defence}. 
The same approach as been applied in NLP models to remove sensitive attributes from the representation for privacy or fairness concern~\cite{li2018towards,elazar2018adversarial,coavoux2018privacy}. 
Another approach for removing sensitive attributes is through directly minimizing the mutual information between the sensitive attributes and the deep representations~\cite{osia2018deep,moyer2018invariant}.
We adopt the adversarial training approach during training embedding models as defense against embedding inversion and sensitive attribute inference attacks.

\paragraphb{Inverting deep representations.} In computer vision community, inverting deep image representation has been studied as a way for understanding and visualizing traditional image feature extractor and deep convolutional neural networks.
Both optimization-based approach~\cite{mahendran2015understanding,ulyanov2018deep} and learning-based approach~\cite{dosovitskiy2016inverting,dosovitskiy2016generating} as been proposed for inverting image representations. 
In contrast, we focus on text domain data which is drastically different than image domain data. Text data are discrete and sparse in nature while images are often considered as continuous and dense.
Our proposed inversion methods are tailored for unsupervised text embedding models trained with recurrent neural networks and Transformers. 

Model inversion attacks~\cite{fredrikson2015model} use gradient based method to reconstruct the input data given a classifier model and a class label. 
The reconstruction is often a class representatives from the training data, e.g. averaged face images of female for a gender classifier~\cite{melis2019exploiting}. 
Embedding inversion, on the other hand, takes the representation vector as input and reconstruct the exact raw input text. 

\paragraphb{Membership inference and memorization.} Membership inference attacks (MIA) have first been studied against black-box supervised classification models~\cite{shokri2017membership}, and later on generative models and language models ~\cite{hayes2019logan,song2019auditing}.
MIA is closely connected to generalization where overfitted models are prone to the attacks~\cite{yeom2018privacy}.
In this work, we extended the study of MIA to unsupervised word and sentence embedding models without a clear notion of generalization for such models.  

It has been shown that deep learning models have a tendency to memorize~\cite{zhang2016understanding}. 
Later work showed that adversaries can extract formatted training text from the output of text generation models~\cite{carlini2019secret}, indicating a real privacy threat caused by memorization. 
In this work, we focus on embedding models where the output is an embedding vector without the possibility of extract training data directly from the output as in text generation models. 
Instead, we demonstrated how to measure the memorization in embeddings through MIA.


\paragraphb{Differential Privacy.} Differentially-private (DP) training of ML models~\cite{abadi2016deep,mcmahan2017learning} involves clipping and adding noise to instance-level gradients and is designed to train a model to prevent it from memorizing training data or being susceptible to MIA. DP, which limits how sensitive the model is to a training example does not provide a defense against attacks that aim to infer sensitive attributes (which is an aggregate property of training data). The noisy training techniques are challenging for models with large parameters on large amounts of data and sensitive to the hyper-parameters~\cite{papernot2020making}. Our embeddings, with over 10 million parameters in the word embedding matrices $\mV$ make DP training and hyper-parameter tuning computationally infeasible. Therefore, we leave it to future work to  explore how to efficiently train embeddings with DP. 


\section{Conclusions}
In this paper, we proposed several attacks against embedding models exploring different aspects of their privacy.
We showed that embedding vectors of sentences can be inverted back to the words in the sentences with high precision and recall, and can also reveal the authorship of the sentences with a few labeled examples.
Embedding models can also leak moderate amount of membership information for infrequent data by using similarity scores from embedding vectors in context.
We finally proposed defenses against the information leakage using adversarial training and partially mitigated the attacks at the cost of minor decrease in utility.

Given their enormous popularity and success, our results strongly motivate the need for caution and further research. Embeddings not only encode useful  semantics of unlabeled data but often sensitive information about input data that might be exfiltrated in various ways. When the inputs are sensitive, embeddings should not be treated as simply ``vectors  of real numbers.''
\newpage



\bibliographystyle{ACM-Reference-Format}
\bibliography{citation.bib}

\end{document}